\documentclass[sigconf]{acmart}
\AtBeginDocument{%
  }

\usepackage{microtype}
\usepackage{tabularx}
\usepackage{graphicx}
\usepackage{xspace}
\usepackage{mdframed}
\usepackage{multirow} 
\usepackage{natbib}

\newcommand{\model}{\textsc{LLMaEL}\xspace}
\newcommand{\insgenel}{\textsc{InsGenEL}\xspace}

\setcopyright{acmlicensed}
\copyrightyear{2025}
\acmYear{2025}
\setcopyright{cc}
\setcctype{by}
\acmConference[CIKM '25]{Proceedings of the 34th ACM International Conference on Information and Knowledge Management}{November 10--14, 2025}{Seoul, Republic of Korea}
\acmBooktitle{Proceedings of the 34th ACM International Conference on Information and Knowledge Management (CIKM '25), November 10--14, 2025, Seoul, Republic of Korea}
\acmDOI{10.1145/3746252.3761156}
\acmISBN{979-8-4007-2040-6/2025/11}

\begin{document}

\title{\model: Large Language Models are Good
Context Augmenters for Entity Linking}

\author{Amy Xin}
\authornote{Equal Contribution.}
\orcid{0009-0001-2404-0475}
\affiliation{%
  \department{Department of Computer Science and Technology, Beijing National Research Center for Information Science and Technology}
  \institution{Tsinghua University}
  \city{Beijing}
  \country{China}
}
\email{xin-x25@mails.tsinghua.edu.cn}

\author{Yunjia Qi}
\authornotemark[1]
\affiliation{%
  \department{Department of Computer Science and Technology, Beijing National Research Center for Information Science and Technology}
  \institution{Tsinghua University}
    \city{Beijing}
  \country{China}
}
\email{qyj23@mails.tsinghua.edu.cn}

\author{Zijun Yao}
\affiliation{%
  \department{Department of Computer Science and Technology, Beijing National Research Center for Information Science and Technology}
  \institution{Tsinghua University}
  \city{Beijing}
  \country{China}
}
\email{yaozj20@mails.tsinghua.edu.cn}

\author{Fangwei Zhu}
\affiliation{%
 \institution{Beijing University}
 \city{Beijing}
 \country{China}}
 \email{zhufangwei2022@stu.pku.edu.cn}

\author{Kaisheng Zeng}
\affiliation{%
  \institution{Tsinghua University}
  \city{Beijing}
  \country{China}
}
\email{zks19@mails.tsinghua.edu.cn}

\author{Bin Xu}
\affiliation{%
  \institution{Tsinghua University}
  \city{Beijing}
  \country{China}
}
\email{xubin@tsinghua.edu.cn}

\author{Lei Hou}
\authornote{Corresponding Authors.}
\affiliation{%
  \department{Department of Computer Science and Technology, Beijing National Research Center for Information Science and Technology}
  \institution{Tsinghua University}
  \city{Beijing}
  \country{China}
}
\email{houlei@tsinghua.edu.cn}

\author{Juanzi Li}
\authornotemark[2]
\affiliation{%
  \department{Department of Computer Science and Technology, Beijing National Research Center for Information Science and Technology}
  \institution{Tsinghua University}
  \city{Beijing}
  \country{China}
}
\email{lijuanzi@tsinghua.edu.cn}

\renewcommand{\shortauthors}{Amy Xin et al.}

\begin{abstract}
Specialized entity linking (EL) models are well-trained at mapping mentions to unique knowledge base (KB) entities according to a given context.
However, specialized EL models struggle to disambiguate long-tail entities due to their limited training data.
Meanwhile, extensively pre-trained large language models (LLMs) possess broader knowledge of uncommon entities.
Yet, with a lack of specialized EL training, LLMs frequently fail to generate accurate KB entity names, limiting their standalone effectiveness in EL.
With the observation that LLMs are more adept at context generation instead of EL execution, 
we introduce LLM-Augmented Entity Linking (\model), 
the first framework to enhance specialized EL models with LLM data augmentation.
\model leverages off-the-shelf, tuning-free LLMs as context augmenters, generating entity descriptions to serve as additional input for specialized EL models.
Experiments show that \model sets new state-of-the-art results across 6 widely adopted EL benchmarks: compared to prior methods that integrate tuning-free LLMs into EL, \model achieves an absolute 8.9\% gain in EL accuracy. 
We release our code and datasets~\footnote{https://github.com/THU-KEG/LLMAEL}.
\end{abstract}

\ccsdesc[300]{Software and its engineering~Entity relationship modeling}
\ccsdesc[500]{Computing methodologies~Natural language generation}

\keywords{Entity Linking, Entity Disambiguation, Large Language Models, Knowledge Graphs, Data Augmentation}


\maketitle

\section{Introduction}

Entity linking (EL) is the task of establishing connections between mentions in textual contexts and unique entities in a knowledge base (KB).
It plays an important role in many applications that require semantic understanding,
including question answering (QA)~\citep{korc,wikiwebqsp,xin2025atomr}, 
dialogue generation~\citep{elfordialogue1,elfordialogue2}, and recommendation systems~\citep{wang2022towards,balloccu2022post}.
Effective entity linking is crucial to the reliable performance of such downstream tasks. 
For example, when a QA system answers the question \textit{``When was Michael Jordan born?''}, the system must correctly disambiguate whether the mention ``Michael Jordan'' refers to the basketball player \textit{Michael Jordan} or the UCB professor \textit{Michael Jordan} in order to output accurate results.

However, entity linking is still a challenging task as it requires two distinct capabilities:
(a) Task Specification, which encompasses a thorough understanding of the entity linking task and the precise requirement for its output format, and 
(b) Entity Knowledge, which involves the possession of substantial knowledge about the target entity.
Trained specifically for entity linking, specialized EL models~\citep{blink,genre,refined} excel in task specification, capable of producing accurate entity names that exactly satisfy the format requirement of the EL task.
Meanwhile, extensively pre-trained large language models (LLMs)~\citep{gpt3,llama2} are natural repositories of expansive world knowledge, possessing vast parametric knowledge pertinent to any given entity.

As LLMs have gained widespread recognition and accessibility in recent years, several prior studies have explored the integration of LLMs into the entity linking task. 
Existing methods can be grouped into two categories: 
(1) \textit{Fine-tuning-based methods}, which involve training LLMs for entity linking through supervised fine-tuning~\citep{insgenel};
(2) \textit{Prompting-based methods}, which employ in-context learning to prompt LLMs for entity linking~\citep{chatel, gemel}.
However, both approaches directly employ LLMs as \textit{EL executors}, leading to drawbacks for each method.
First, fine-tuning-based methods, despite yielding promising results, require significant training expenses to tune LLMs' massive parameters. 
Conversely, prompting-based methods are more cost-efficient, but yield less accurate results due to LLMs' limitations in EL task specification.
Hence, we ask the following question: \textit{How can we integrate LLMs into the entity linking task to effectively balance computational cost and performance?}

Motivated by this challenge, we introduce LLM-Augmented Entity Linking (\model), 
the first entity linking framework to employ LLMs as data augmenters for specialized EL models.
Unlike previous approaches that directly utilize LLMs as \textit{EL executors}, \model employs LLMs as \textit{EL context augmenters}, providing specialized EL models with supplemental context to aid entity linking execution.
\model effectively addresses the limitations of previous works, particularly the trade-off between computational cost and performance.
In terms of cost, \model interacts with LLMs via prompting, offering a lightweight solution free from expensive LLM fine-tuning. 
In terms of performance, \model achieves promising results by leveraging specialized EL models for final entity linking execution.
Overall, \model successfully combines the extensive entity knowledge
of LLMs with the
task-specific capabilities of specialized EL models, delivering an LLM-enhanced EL solution that effectively balances computational efficiency and performance accuracy.

Our paper's contributions are as follows:

\begin{enumerate}
\item We propose \model, the first entity linking framework that leverages LLMs as context augmenters for specialized EL models. 
We show that \model is a flexible framework with multiple advanced configurations that further enhance its performance.
\item 
\model yields promising results: 
compared to prior methods that integrate tuning-free LLMs into the EL task, \model yields absolute accuracy gains as high as $8.9\%$;
compared to only using specialized EL models, \model achieves absolute average accuracy improvements of $1.2\%$.

\item We further explore alternative means of integrating tuning-free LLMs into the entity link process, and demonstrate that \model is the most effective approach.
\end{enumerate}

\section{Preliminaries and Related Work}
\label{sec:related_work}


\subsection{Task Definition}

Entity Linking (EL) is the task of mapping mentions from a given context to KB entities. 
Formally, the knowledge base $G$ consists of the set of entities $\{e\}$ that are unique objects in the real world.
The input of entity linking is a textual context $c$, embedded with multiple entity mentions, denoted as $c = \dots {t}_1 || {m}_1 || {t}_2 || {m}_2 || {t}_3 \dots$, where $t_i$ are textual spans and $m_i$ are entity mentions.
The goal of entity linking is to obtain a correct list of mention-entity pairs $\{(m_i, e_i)\}_{i \in [1,k]}$.

\subsection{Related Work}

\paragraph{Specialized Entity Linking Models.}

It has been a long-standing goal to develop models specialized for entity linking.
The most widely adopted solution to build specialized EL models is a two-stage architecture~\citep{sevgili2022neural}, 
which divides EL into two sequential phases: entity candidate generation and entity re-ranking.
Most models approach the entity candidate generation phase as a retrieval problem, aligning mentions to entities according to BM25 ranking\citep{retrieval2} or embedding similarity scores\citep{blink, retrieval3}. 
With the development of generative language models, 
it has also become possible to treat entity candidate generation as a text generation task~\citep{genre}, training the model to generate unique entity names in the knowledge base directly based on the contextual information. 
Furthermore, most recent works prove that concept information about mentions is useful for EL, thus fine-grained entity typing is also integrated as part of the pipeline and has been applied to
numerous models \citep{refined, entity_typing2}.
This suggests that augmenting mentions with additional information facilitates the entity linking process, leaving space for further research on EL context augmentation.

\paragraph{LLMs as Executors for Downstream Tasks.}

In-context learning, or few-shot prompting, is a prevalent strategy that enables LLMs to execute downstream tasks without the need of LLM fine-tuning. 
With the outstanding accomplishments of LLMs like GPT-3 \citep{gpt3} and LLaMA2 \citep{llama2}, LLM in-context learning has achieved impressive results in downstream task execution, such as question answering, summarization, and machine translation.
However, research shows that LLM in-context learning still falls short when executing \textit{specification-heavy tasks} \citep{peng2023does}, which are tasks with complicated and extensive task specifications (e.g., entity linking).
For these tasks, LLM in-context learning yields significantly inferior results compared to specialized models.
Therefore, relying on LLM in-context learning for direct entity linking execution is not the optimal approach.
This underscores the need for more effective means to integrate LLMs into the entity linking process.

\paragraph{LLMs as Context Augmenters for Downstream Tasks.}
\label{par:llms_for_downstream_tasks}
LLMs are primarily designed for text generation, which is their strongest advantage. 
Numerous studies have shown that LLM-generated contexts present outstanding qualities, outperforming contexts obtained from information retrieval methods \citep{generate, beyond}. 
Furthermore, compared to retrieved contexts, LLM-generated contexts contribute to better downstream task performance \citep{beyond}. 
With such insights, a bright solution is to leverage LLMs as \textit{context augmenters} instead of task executors: to generate supplemental context for specialized models as additional input.
For instance, \citet{liu2021generated}
demonstrate how LLM context augmentation can enhance commonsense reasoning, achieving state-of-the-art results on multiple reasoning tasks.
Similarly, \citet{balkus2022improving}
show that GPT-3 augmented data can improve specialized text classification models, resulting in higher consistent accuracy on unseen examples. 
However, there have been no attempts on employing LLM context augmentation to enhance specialized entity linking models.

\paragraph{LLMs for Entity Linking.}
\label{par:llms_for_el}
Several prior works have integrated LLMs into the entity linking task.
Yet, all works employ LLMs as direct task executors for entity linking.
The methodologies of these works can be categorized into two lines of approaches:

\textbf{Fine-tuning-based methods.}  
This approach leverages LLMs by directly fine-tuning them specifically for the entity linking task.
At test time, the fine-tuned LLM serves as the entity linking model to directly generate entity predictions.  
A representative example is \insgenel \citep{insgenel}, which introduces a sequence-to-sequence training objective for EL and applies instruction fine-tuning to enable LLMs to execute the EL task. 
However, while effective, the fine-tuning approach is computationally expensive, making it a costly solution.  

\textbf{Prompting-based methods.}  
In contrast to fine-tuning, prompting-based methods employ LLMs as entity linkers through in-context learning. 
These methods design carefully crafted prompts for EL, which are then inputted into the LLM to generate entity predictions.  
For instance, ChatEL \citep{chatel} presents a three-step framework to prompt LLMs for EL.
Similarly, GEMEL \citep{gemel} applies prompting-based techniques to multimodal EL tasks and incorporates a constrained beam search strategy to refine entity generation.  
However, these methods face challenges due to the inherent limitations of LLMs in accurately generating unique entity names, as they lack specialized training on the task. 
Consequently, prompting-based methods often yield inferior results compared to specialized EL models like ReFinED \citep{refined} and GENRE \citep{genre}, which are both more effective and cost-efficient.

\paragraph{Comparing \model to Previous Works.}
\label{par:compare_to_previous_works}

\model is the first approach to enhance entity linking by employing LLM context augmentation over specialized EL models.
\model could be compared to two categories of existing work: (1) methods that use LLM context augmentation to enhance downstream tasks, and (2) methods that employ LLMs for entity linking.

For category (1), as detailed in the \textit{LLMs as Context Augmenters for Downstream Tasks} paragraph (\ref{par:llms_for_downstream_tasks}), there are methods that successfully employ LLM context augmentation to enhance specialized models in other tasks, such as commonsense reasoning and text classification. 
However, there has been no exploration into how LLM context augmentation can be used to enhance specialized entity linking models, and \model is the first attempt.

Similarly, for category (2), as detailed in the \textit{LLMs for Entity Linking} paragraph (\ref{par:llms_for_el}), there are works that leverage LLMs for entity linking, namely both \textit{fine-tuning-based methods} and \textit{prompting-based-methods}.
However, all these methods directly employ LLMs as EL executors, which leads to either large training costs in the fine-tuning scenario or inferior performance in the tuning-free scenario.
In contrast, \model is the first method to employ tuning-free LLMs as \textit{EL context augmenters} for specialized EL models, effectively addressing the cost-performance trade-off.

To sum up, \model stands as a novel approach that explores the effective collaboration of general LLMs with specialized models under the task of entity linking, combining the strengths of both model types at a low cost.

\section{Methodology}
\label{sec:methodology}

\begin{figure*}[!ht]
    \centering
    \includegraphics[width=0.95\linewidth]{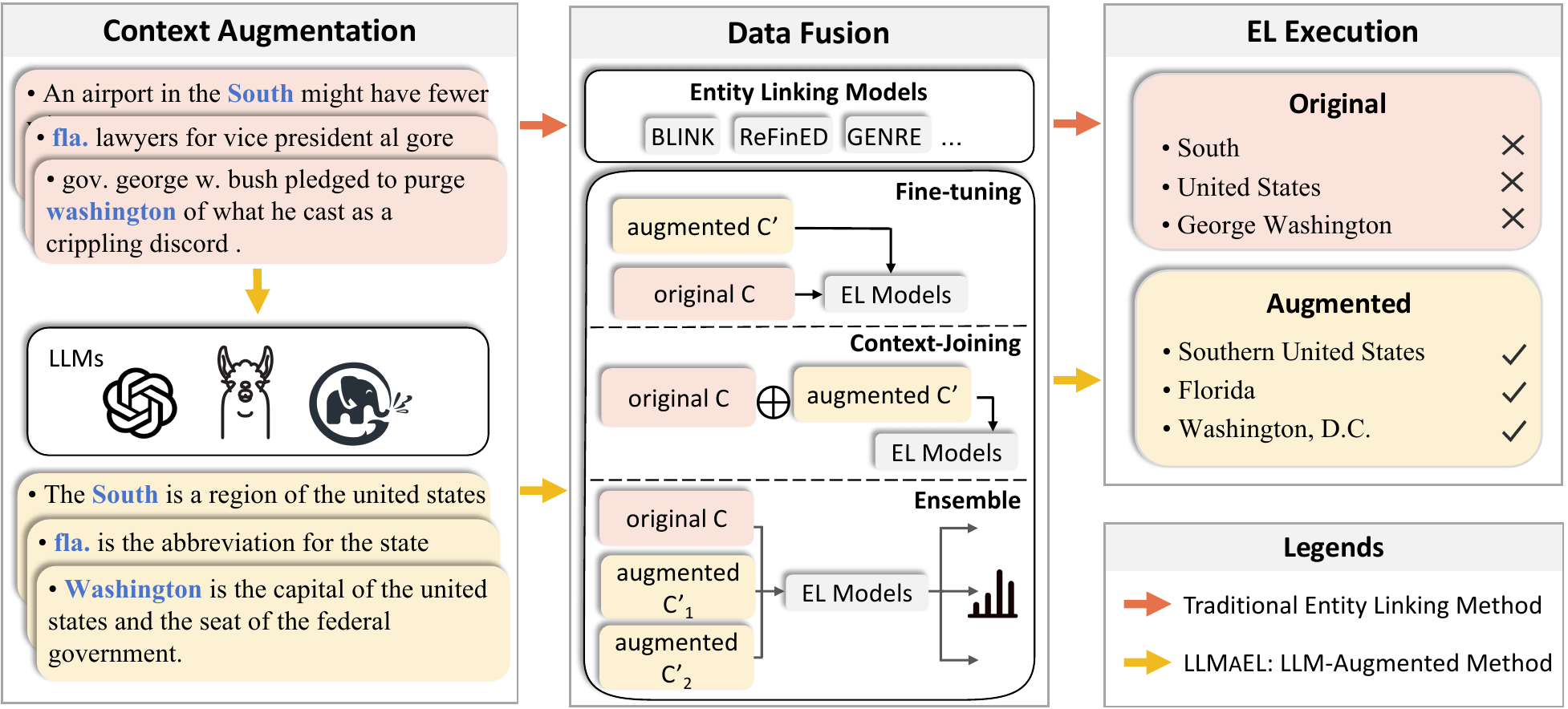}
    \caption{Proposed methodology architecture.}
    \label{fig:pipeline}
\end{figure*}

\model is the first plug-and-play framework to employ LLMs as context augmenters for specialized EL models.
Figure \ref{fig:pipeline} illustrates the overall system architecture of \model. 
We mark the traditional entity linking process in pink and our method in yellow.
Mentions that need to be executed by entity linking are marked with blue.
As presented by Figure \ref{fig:pipeline}, \model mainly includes three building blocks. 
(1) \textbf{Context augmentation} is the most basic element for \model, which elicits LLMs to generate enriched context with more information for entity linking.
(2) \textbf{Data fusion} designs multiple strategies to integrate the LLM-generated content with the original context, aiming to improve diverse off-the-shelf EL models.
(3) \textbf{EL execution} finally conducts the entity linking task.

\subsection{Context Augmentation}
\label{context-augmentation}

In the Context Augmentation stage, we employ tuning-free LLMs to generate supplementary context for a given mention. 
Formally, we denote LLM context augmentation as the function
$$
c' = \text{LLM}(p, c, m_i)
$$
where $p$ is a specially designed prompt to instruct LLMs for context generation, $c$ is the original context for the $i^\text{th}$ mention $m_i$, and $c'$ is the augmented context generated by the LLM. 

We implement LLM context augmentation through in-context learning~\citep{gpt3} with a three-shot prompt.
Our prompt comprises two main parts: (1) task specification instructions on context augmentation, and (2) in-prompt demonstrations.
For task specification, we use the following template to prompt LLMs:
\vspace{3.5pt}
\begin{mdframed}
\begin{verbatim}
 Consider the following text.
 Text: [CONTEXT]
 Please provide me more descriptive information 
 about [MENTION] from the text above. 
 Make sure to include [MENTION] in your description.
\end{verbatim}
\end{mdframed}
\vspace{3.5pt}
where $\texttt{[CONTEXT]}$ and $\texttt{[MENTION]}$ are placeholders to be filled before feeding into LLMs.
It is worth noting that our instruction requires LLMs to recite the entity again in the generated context, which provides flexibility for later data fusion.
For in-prompt demonstrations, we bootstrap examples via zero-shot prompting.
To ensure the quality of these demonstrations, we first generate a sufficiently large number of LLM responses. 
via zero-shot prompting, and then manually filter out high-quality samples. 
The final prompt's exemplars are then selected from this high-quality pool.


\subsection{Data Fusion}
\label{sec:enhancement_options}

The Data Fusion stage designs multiple strategies to integrate LLM-augmented context $c'$ into the EL process.
In particular, \textit{context-joining strategies} directly augment $c'$ into the original context $c$;
\textit{EL model fine-tuning} infuses the knowledge in $c'$ into the EL model's parameters via lightweight fine-tuning; and \textit{ensemble} combines multiple LLM-augmented contexts through output majority voting.

\begin{table}
\centering
\caption{\label{join-strategies} Five possible context-joining strategies for \model.}
\scalebox{1.1}{
\begin{tabular}{c|l|l}
\toprule
\textbf{ID} & \textbf{Context Order} & \textbf{Mention Offset}\\
\midrule
$0$ & LLM-only & LLM \\
$1$ & LLM + Original & LLM \\
$2$ & LLM + Original & Original \\
$3$ & Original + LLM & LLM \\
$4$ & Original + LLM & Original \\
\bottomrule
\end{tabular}
}
\end{table}

\paragraph{Context-Joining Strategies.}

A critical consideration of \model's implementation is determining how to integrate the LLM-augmented context \( c' \) with the original context \( c \) from the text source. 
The most straightforward approach is to concatenate \( c' \) and \( c \) into a whole passage, then input the combined context into off-the-shelf EL models. 
This approach raises two key design questions:  

\begin{enumerate}
    \item \textbf{Context Order}: In what sequence should \( c \) and \( c' \) be concatenated?
    \item \textbf{Mention Offset}: Since the mention appears in both \( c \) and \( c' \), which mention span should be selected as the target span for the EL models?
\end{enumerate}

To address these questions, \model considers five potential strategies for joining the original and LLM-generated contexts, as outlined in Table~\ref{join-strategies}. These strategies enumerate different arrangements of \( c \) and \( c' \) in the final augmented context provided to the EL model. In Table~\ref{join-strategies}, \textit{Context Order} specifies the sequential arrangement of the original and LLM-generated contexts, while \textit{Mention Offset} determines which context's mention span is used by the EL model.  
Specifically:
\begin{itemize}
    \item \textbf{Strategy 0} substitutes \( c' \) directly for \( c \), discarding the original context.
    \item \textbf{Strategies 1–4} explore all four possible combinations of context order and mention offset.
\end{itemize}

Our experiments reveal that the optimal joining strategy varies across different EL models. Consequently, we treat the choice of context-joining strategy as a tunable hyperparameter, enabling users to adapt \model to various application settings for optimal performance.

\paragraph{EL Model Fine-tuning.}

A key challenge for \model is that the style and distribution of the augmented EL context may differ from what EL models are accustomed to, potentially resulting in suboptimal performance. 
To address this mismatch between the text distributions familiar to EL models and the LLM-augmented contexts, we explore the option of fine-tuning existing EL models.
Specifically, we first perform LLM context augmentation on existing EL training datasets as described in~\ref{context-augmentation}, where an LLM generates mention descriptions for each entry of the training set. 
Then, the generated descriptions are incorporated into the training set using the optimal context-joining strategy of the selected EL model. 
Finally, we fine-tune the EL model on this augmented training set to improve its compatibility with the LLM-augmented contexts.

\paragraph{Ensemble.}

Inaccuracies in \model's performance may occur when the LLM generates incorrect mention descriptions, misguiding the EL model to output an incorrect entity.
Hence, we also expand \model with ensemble techniques, attempting to improve \model's robustness through diversified sampling. 
We sample mention descriptions across multiple LLMs and evaluate the diversified samples through both hard-voting and soft-voting classifier methods. 

\subsection{EL Execution}
In the final stage of EL execution, the 
specialized EL model is 
employed to output the KB entity name.
Compared to directly tasking LLMs to perform entity linking, \model ensures EL task specification using 
specialized EL models, while augmenting 
the EL models with expansive
entity knowledge from the LLM-generated context.

\section{Experimental Setup}
\label{sec:experiment_setup}




\paragraph{Backbone Models.}
For our main experiments, we use Llama-3-70b-Instruct as the backbone LLM.
We select Llama-3-70b-Instruct for two main reasons. 
First, as a context augmentation framework, \model aims to achieve robust performance with open-source LLMs. By leveraging such an open-source LLM, we prioritize contributing to the research community with a solution that is both openly accessible and cost-effective.
Second, \model requires an LLM with exceptional text generation capabilities to meet the demands of EL context augmentation. 
Llama-3-70b-Instruct currently stands as one of the best-performing open-source LLMs for text generation, making it a natural fit for our experiments.
Additionally, our experiments reveal that Llama-3-70b-Instruct delivers comparable, and in some cases superior, performance to proprietary LLMs when applied to \model, which will be detailed in Section \ref{sec:choosing_llms}.
For all experiments, we set the number of LLM maximum output tokens to 150.
We set the temperature of Llama-3-70b-Instruct to 0.01, and adapt default values for all other hyperparameters.

As \model is a plug-and-play framework for any specialized EL models, we evaluate three most widely adopted EL models as our backbones: BLINK \citep{blink}, a classical bi-encoder cross-encoder EL model; GENRE \citep{genre}, an autoregressive generative EL model; and ReFinED \citep{refined}, an EL model enhanced with entity typing and descriptions. 
All three models are state-of-the-art for their respective model architectures. 
Specifically, we use the full BLINK model\footnote{BLINK's full cross-encoder model}, the fairseq-AIDA GENRE model\footnote{The GENRE model developed using the fairseq toolkit and officially fine-tuned on AIDA-YAGO2}, and the AIDA ReFinED model\footnote{The ReFinED model officially fine-tuned on AIDA-YAGO2}.
To ensure a fair assessment across EL models, we evaluate GENRE without the pregenerated candidate set.

\paragraph{Datasets.}

We evaluate \model on six widely adopted entity linking (EL) datasets: AIDA-YAGO2 \citep{aida}, MSNBC \citep{msnbc}, AQUAINT \citep{aquaint}, ACE2004 \citep{ace2004}, WNED-CWEB \citep{cweb}, and WNED-WIKI \citep{wiki}. 
These datasets are selected for their popularity and recognition in the EL field, as well as their ability to cover diverse domains, including news articles, web pages, and encyclopedia corpora. 






For each of our three backbone EL models, we first download their official training and testing splits from their respective github repositories\footnote{https://github.com/facebookresearch/BLINK}\footnote{https://github.com/facebookresearch/GENRE}\footnote{https://github.com/amazon-science/ReFinED}. 
These repositories provide standard versions of the datasets required for both the evaluation and training of \model, including the test splits for MSNBC, AQUAINT, ACE2004, WNED-CWEB, WNED-WIKI, and the train, dev, and test splits for AIDA-YAGO2.
We use our backbone LLM to generate mention descriptions for all eight data splits for the ReFinED model, and just the six test splits for the other two EL models.

\paragraph{Baselines.}

We compare \model with two categories of baselines: 
(1) \textit{LLMs for EL}.
First, we evaluate a naive \textit{LLM only} baseline that prompts an LLM to execute the EL task through in-context learning.
Specifically, we prompt Llama-3-70b-Instruct with three-shot examples of mention-context pairs and gold entity names, and demand it to output the entity name for the current test case.
Second, we compare with ChatEL~\citep{chatel}, the state-of-the-art prompting-based framework that uses tuning-free LLMs for entity linking. 
(2) \textit{Specialized EL models}.
We also compare with each of our three backbone EL models in a standalone setting without any LLM data augmentation.

\paragraph{Evaluation Metrics.}

\begin{table*}[!t]
  \centering
  \caption{\label{ED Results}Main results of \model across six EL benchmarks.} 
  \scalebox{1.1}{
  \begin{tabular}{lcccccc|c}
    \toprule
    \textbf{Method} & \textbf{AIDA} & \textbf{MSNBC} & \textbf{AQUA} & \textbf{ACE04} & \textbf{CWEB} & \textbf{WIKI} & \textbf{AVG}\\
    \midrule
    LLM only & $78.37$ & $80.49$ & $73.18$ & $83.27$ & $65.34$ & $64.44$ & $74.18$ \\
    ChatEL~\citep{chatel} & $81.20$ & $83.50$ & $75.70$ & $85.60$ & $66.20$ & $74.50$ & $77.78$ \\
    \midrule
    BLINK only & $82.01$ & $86.23$ & $85.16$ & $86.01$ & $69.11$ & $81.11$ & $81.61$ \\
    GENRE only & $87.92$ & $83.54$ & $84.32$ & $84.82$ & $68.75$ & $83.02$ & $82.06$ \\
    ReFinED only & $92.25$ & $\underline{87.10}$ & $87.53$ & $\underline{87.75}$ & $72.96$ & $85.18$ & $85.46$ \\
    \midrule
    \model $\times$ BLINK & $81.94$ & $86.56$ & $85.16$ & $86.01$ & $69.17$ & $81.14$ & $81.66$ \\
    \model $\times$ GENRE & $88.27$ & $85.67$ & $85.14$ & $85.21$ & $70.67$ & $82.95$ & $82.99$ \\
    \model $\times$ ReFinED & $\textbf{92.38}$ & $86.94$ & $\underline{88.09}$ & $\textbf{88.14}$ & $\underline{73.16}$ & $\underline{85.90}$ & $\underline{85.76}$\\
    \midrule
    \model $\times$ $\text{ReFinED}_{\text{FT}}$ & $\underline{92.34}$ & $\textbf{88.79}$ & $\textbf{89.06}$ & $\textbf{88.14}$ & $\textbf{75.07}$ & $\textbf{86.62}$ & $\textbf{86.67}$ \\
    \bottomrule
  \end{tabular}
  }
\end{table*}

We use \textit{disambiguation accuracy} as our evaluation metric, defined as the ratio of correctly linked entities to the total number of entities in the dataset:

\[
\text{Accuracy} = \frac{\text{\# Correctly linked entities}}{\text{Total \# of entities in dataset}}
\]


For experiments involving an EL model, we first augment the model's official datasets with LLM-generated context. The augmented datasets are then evaluated using the model's official testing scripts to ensure consistency with prior benchmarks.
For LLM-only baseline experiments that do not incorporate EL models, we manually constructed an evaluation script to compute the LLM's disambiguation accuracy, where we use the \textit{Exact Match (EM)} measure
defined below:

\[
\text{Exact Match (EM)} = 
\begin{cases} 
1 & \text{if } \hat{y} = y \\
0 & \text{otherwise}
\end{cases}
\]

Where $\hat{y}$ is the LLM's predicted entity name and $y$ is the ground truth entity name. 
The measure only outputs $1$ if $\hat{y}$ is exactly equal to $y$, character by character.
Then, we divide the number of exact match entities by the total number of entities in the dataset to obtain the final disambiguation accuracy.

\paragraph{Selecting a Unified Context-Joining Strategy.}

We use the dev split of AIDA-YAGO2 to determine the optimal context-joining strategy. 
In our main experiments, we consistently report results obtained by Strategy 4 across all three EL models, as it achieves the highest average accuracy for each model on the dev set, ensuring a fair comparison. 
We hypothesize that Strategy 4 is superior because most EL models are more accustomed to the distribution of original contexts, thus yielding better performance when LLM-generated contexts are appended strictly after the original contexts.

We conduct context-joining strategy experimentation with the following implementation details.
For joining strategies that incorporate context combination, contexts are merged using a newline symbol ``\textbackslash n''. 
In the case of BLINK and ReFinED, contexts are fully combined in the specified order, without any truncation applied. 
In the case of GENRE, the contexts are first fully combined and then trimmed to the model's maximum input sequence length.

\paragraph{Fine-tuning.}

We select our best-performing EL model ReFinED for model fine-tuning.
We use the train and dev splits from the AIDA-YAGO2 dataset as our training and evaluation data. 
To avoid model over-fitting on AIDA-YAGO2, we leverage ReFinED's wikipedia model
for fine-tuning. 
Specifically, we first employ Llama-3-70b-Instruct to augment the datasets under the model's optimal context-joining strategy, then apply the augmented datasets for EL model fine-tuning.

\section{Results and Discussions}
\label{sec:results}

\subsection{Main Results}\label{sec:main_exp}

The results of our main experiments are presented in Table~\ref{ED Results}. 
Table~\ref{ED Results}'s results can be divided into four horizontal sections, ordered from top to bottom.
The first section presents results of baselines that only leverage tuning-free LLMs.
The second section presents results of standalone EL models.
The third section presents results of the vanilla implementation of \model, representing a straightforward context augmentation approach without any EL model fine-tuning.
The final section reports the results of \model with a fine-tuned ReFinED model, denoted as $\text{ReFinED}_{\text{FT}}$.
For each dataset, the highest disambiguation accuracy score is highlighted in bold, while the second-highest score is underlined.
The results of ChatEL are taken from their official github repository \footnote{\url{https://github.com/yifding/In_Context_EL}}, while all other results are produced by ourselves.

The vanilla implementation of \model improves the average accuracy score of the six datasets across all three EL models.
Vanilla \model enhances performance on at least five datasets for each EL model, with \model $\times$ GENRE demonstrating an average improvement of $0.93\%$ over its EL model backbone.
It is also important to note that, while both incorporating LLMs under a completely tuning-free scenario, vanilla \model yields significantly better results than the ChatEL baseline (7.68\% higher accuracy with \model $\times$ ReFinED), justifying the effectiveness of our method design.

The fine-tuned \model $\times$ $\text{ReFinED}_{\text{FT}}$ consistently delivers new state-of-the-art results across all six datasets, offering a $1.21\%$ accuracy improvement over the original ReFinED backbone and a $0.91\%$ increase compared to the vanilla \model $\times$ ReFinED. 
This supports our hypothesis that fine-tuning further enhances performance by better aligning EL models with the token distributions of LLM-augmented contexts.
Compared to ChatEL, \model $\times$ $\text{ReFinED}_{\text{FT}}$ further yields a performance gain of $8.89\%$.

Two datasets in particular are worth specific discussion in the \model $\times$ $\text{ReFinED}_{\text{FT}}$ configuration.
First, on the AIDA test set, there is a slight performance reduction from $92.38\%$ to $92.34\%$. 
We hypothesize that this is due to potential overfitting to the AIDA dataset distribution during fine-tuning with $\text{ReFinED}_{\text{FT}}$, resulting in a minor performance drop.
Secondly, ACE2004 results appear capped at $88.14\%$, both with and without fine-tuning. 
This suggests that $88.14\%$ may be the maximum achievable accuracy for the ReFinED model on the ACE2004 test set, likely due to inherent limitations of the dataset and the model.
On all other datasets, $\text{ReFinED}_{\text{FT}}$ significantly boosts performance.

\subsection{Ablation Studies} \label{sec:ablations}

In this section, we conduct ablation studies to test the robustness of \model and to experiment with additional strategies that further enhance performance.
In Section \ref{sec:choosing_llms}, we employ other state-of-the-art LLMs on our method to demonstrate \model's robustness across LLM choices, as well as to justify our choice of Llama-3-70b-Instruct as the primary LLM backbone.
In Sections \ref{sec:join-strat} and \ref{sec:majority_voting}, we explore additional strategies that further boosts \model's performance.

\subsubsection{Choosing Among LLMs\label{sec:choosing_llms}}

\begin{table*}
\centering
\caption{\label{LLMs and mv} Results of \model across different backbone LLM choices. }
\scalebox{1.1}{
\begin{tabular}{l|l|l|lll|l}
\toprule
  & \textbf{EL Model}     & \textbf{\model LLM backbone(s)}     & \textbf{MSNBC}  & \textbf{AQUA}   & \textbf{WIKI}   & \textbf{AVG}                  \\ \midrule
\multirow{8}{*}{\rotatebox{90}{Single}} & \multirow{4}{*}{ReFinED}    & -& $87.10$  & $87.53$  & $85.18$  & $86.60$                \\
  &       & Llama-3-70b-Instruct  &  $86.94$ & $88.09$ & $85.90$ & $86.98$ \\ 
  &       & GPT-3.5-Turbo-Instruct & $86.94$  & $88.23$  & $85.60$  & $86.92$                \\
  &       & GLM-4& 86.94  & $87.95$  & $85.75$  & $86.88$                \\ \cmidrule{2-7} 
  & \multirow{4}{*}{ReFinED$_{\text{FT}}$} & -& $89.40$& $89.20$& $85.93$& $88.18$ \\
  &       & Llama-3-70b-Instruct  &$88.79$	&$89.06$	&$86.62$	&$88.16$ \\  
  &       & GPT-3.5-Turbo-Instruct  &$89.40$	&$89.47$	&$86.28$	&$88.38$ \\
  &       & GLM-4     &$89.40$	 &$89.20$ &	$86.24$ &	$88.28$ \\  
  \midrule
\multirow{4}{*}{\rotatebox{90}{Multi}}  & \multirow{2}{*}{ReFinED}    & Hard-voting ensemble &$86.94$	 &$87.95$	 &$85.90$	 &$86.93$ \\
  &       & Soft-voting ensemble & $86.94$	 &$87.95$	 &$85.75$	 &$86.88$ \\
  \cmidrule{2-7} 
  & \multirow{2}{*}{ReFinED$_{\text{FT}}$} & Hard-voting ensemble &$89.09$	& $89.47$	&$86.62$	&$88.39$ \\
  &       & Soft-voting ensemble & $89.40$	& $89.34$	&$86.25$	&$88.33$ \\
\bottomrule
\end{tabular}
}
\end{table*}

As \model is adaptable to any LLM, in this section, we investigate \model's performance across different backbone LLM choices.
We compare our primary backbone LLM Llama-3-70b-Instruct to two other widely recognized proprietary LLMs, 
namely GPT-3.5-Turbo-Instruct and GLM-4.
We set the temperature of Llama-3-70b-Instruct and GLM-4 to 0.01 and the temperature of GPT-3.5-Turbo-Instruct to 0, and adapt default values for all other hyperparameters.

The upper half labeled ``Single'' of Table \ref{LLMs and mv} presents the results of backbone LLM ablation.
Again, \model $\times$ ReFinED presents our vanilla context augmentation approach, while $\text{ReFinED}_{\text{FT}}$ presents \model with our custom fine-tuned ReFinED EL model. 
For the vanilla \model $\times$ ReFinED, all three LLMs demonstrate an average performance enhancement. 
Among them, Llama-3-70b-Instruct yields the most significant overall improvement, achieving an average accuracy of $86.98\%$. 
GPT-3.5-Turbo-Instruct and GLM-4 demonstrate comparable performance, yielding average improvements of at least $0.28\%$ .
The fine-tuned $\text{ReFinED}_{\text{FT}}$ also demonstrates excellent compatibility across various LLMs.
With GPT-3.5-Turbo-Instruct, $\text{ReFinED}_{\text{FT}}$ achieves an average performance of $88.38\%$, representing a $1.78\%$ improvement over the original context and model.
It is also noteworthy that the performance of $\text{ReFinED}_{\text{FT}}$ also shows a significant enhancement when applied to the original contexts, registering an average performance improvement of $1.58\%$.

Overall, different LLM backbones contribute only minor performance variations to our method, with a maximum difference of $0.38\%$ in the vanilla setting and $0.22\%$ in the fine-tuned setting. 
This leads to two key conclusions. 
First, it confirms that \model is a robust framework that delivers consistent performance across a variety of LLMs. 
This allows us considerable flexibility in choosing among various open-source and proprietary LLMs while maintaining the framework's promising performance. 
Second, the fact that Llama-3-70b-Instruct achieves the best performance in the tuning-free setting further validates our primary backbone LLM choice. This highlights that our system performs best with open-source LLMs, offering an effective, accessible, and cost-efficient solution.

\subsubsection{Model-Specific Context-Joining Strategies} \label{sec:join-strat}

\begin{table}
\centering
\caption{\label{join_strategy} Results of vanilla \model implemented with BLINK under different context-joining strategies.}
\scalebox{1.1}{
\begin{tabular}{ll|c}
\toprule
\textbf{Method} & \textbf{ID} & \textbf{AVG acc.}\\
\midrule
BLINK only & - & $81.61$ \\
\model $\times$ BLINK & 4 & $81.66$ \\
\model $\times$ BLINK & 1* & \textbf{$84.70$} \\
\bottomrule
\end{tabular}
}
\end{table}

For our main experiments, we used the development set of the AIDA benchmark to select a unified context-joining strategy.
However, during our development set experiment, we observed that different EL models actually show optimal performance with different context-joining strategies.
This is because our three selected EL models each possess different model architectures, leading each model to prefer different data distributions during inference.

Namely, BLINK's optimal context-joining strategy significantly diverges from ReFinED and GENRE.
BLINK's optimal context-joining strategy is strategy 1, as presented in Table \ref{ED Results}.
The table's \textit{AVG acc.} column presents the unweighted macro average disambiguation accuracy score of vanilla \model $\times$ BLINK over all 6 test sets. 
Results show that adopting this model-specific optimal strategy for BLINK leads to a significant performance enhancement of $3.04\%$ in average accuracy. 
Intriguingly, BLINK's optimal test-time strategy (strategy $1$) has completely opposite parameters as unified strategy $4$.
We hypothesize that the reliance on AIDA-dev for selecting the optimal joining strategy might be a contributing factor. 
Given that BLINK, unlike the other two EL models, is not fine-tuned on the AIDA dataset, it may not resonate well with the textual distributions of AIDA datasets. 
Consequently, BLINK's performance on the AIDA-dev dataset does not accurately reflect its true preferences and capabilities.

\subsubsection{Ensemble\label{sec:majority_voting}}

\begin{figure}[!t]
    \centering
    \includegraphics[width=1.0\linewidth]{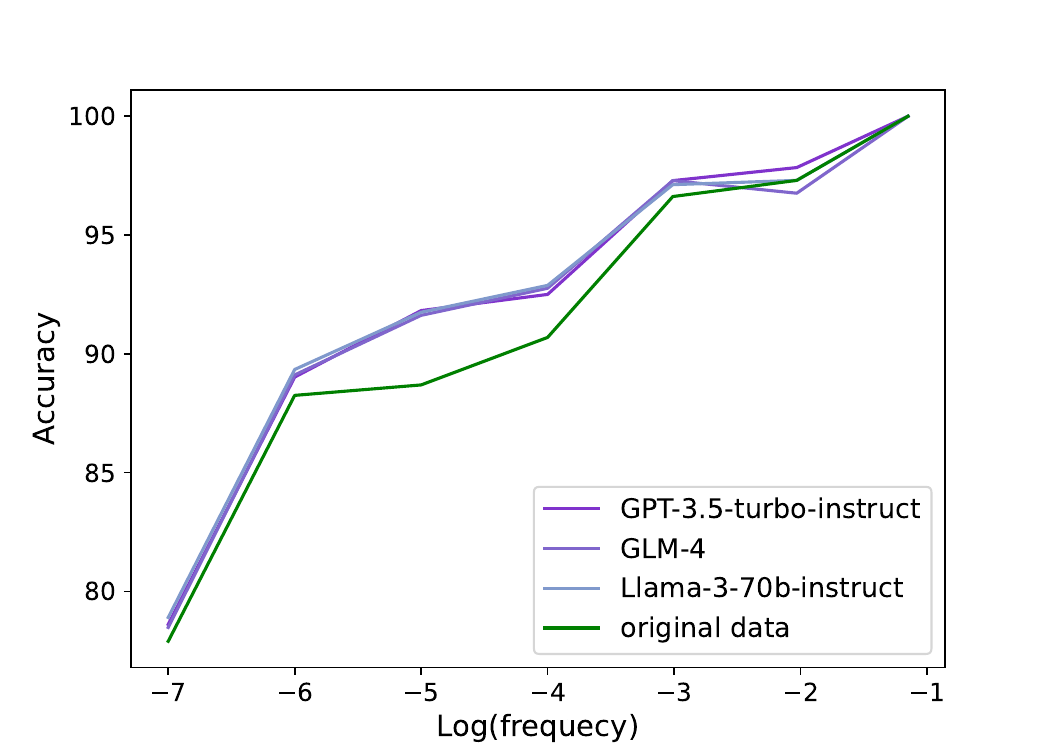}
    \caption{
        Results of \model across entities of different frequencies.
    }
    \label{fig:frequency_bin}
\end{figure}

We use both hard-voting and soft-voting classifiers to perform ensemble. 
The hard-voting classifier is executed by selecting the most frequent outcome among multiple independently-generated results. 
In instances where multiple results share an equivalent frequency, the result with the highest probability level is selected.
Conversely, the soft-voting classifier selects the final answer by aggregating the probabilities of all outcomes.

The \textit{Multi} half of Table~\ref{LLMs and mv} illustrates our ensemble results. 
Both ReFinED and $\text{ReFinED}_{\text{FT}}$ are improved by the implementation of ensemble techniques.
For 
ReFinED, ensemble using the hard-voting classifier achieves the highest average accuracy of $86.93\%$. 
This accuracy score is higher than the score obtained by the soft-voting classifier.
This is because the hard-voting classifier is particularly effective when the performance of individual models is diverse.
For most datasets, the original ReFinED model yields results that are apparently different to the other $3$ LLM-enhanced models, contributing to the diversity of model performance. 
Meanwhile, when the performance of single models is relatively uniform, both ensemble methods—hard and soft-voting classifiers—exhibit comparable effectiveness. 
This phenomenon is evident in the performance outcomes of the $\text{ReFinED}_{\text{FT}}$, where the hard and soft-voting classifiers present equal average accuracies. 
Moreover, the hard-voting classifier under $\text{ReFinED}_{\text{FT}}$ also achieved the best average performance of $88.39\%$.

\subsection{Discussions}

\begin{figure*}[!t]
    \centering
    \includegraphics[width=0.8\linewidth]{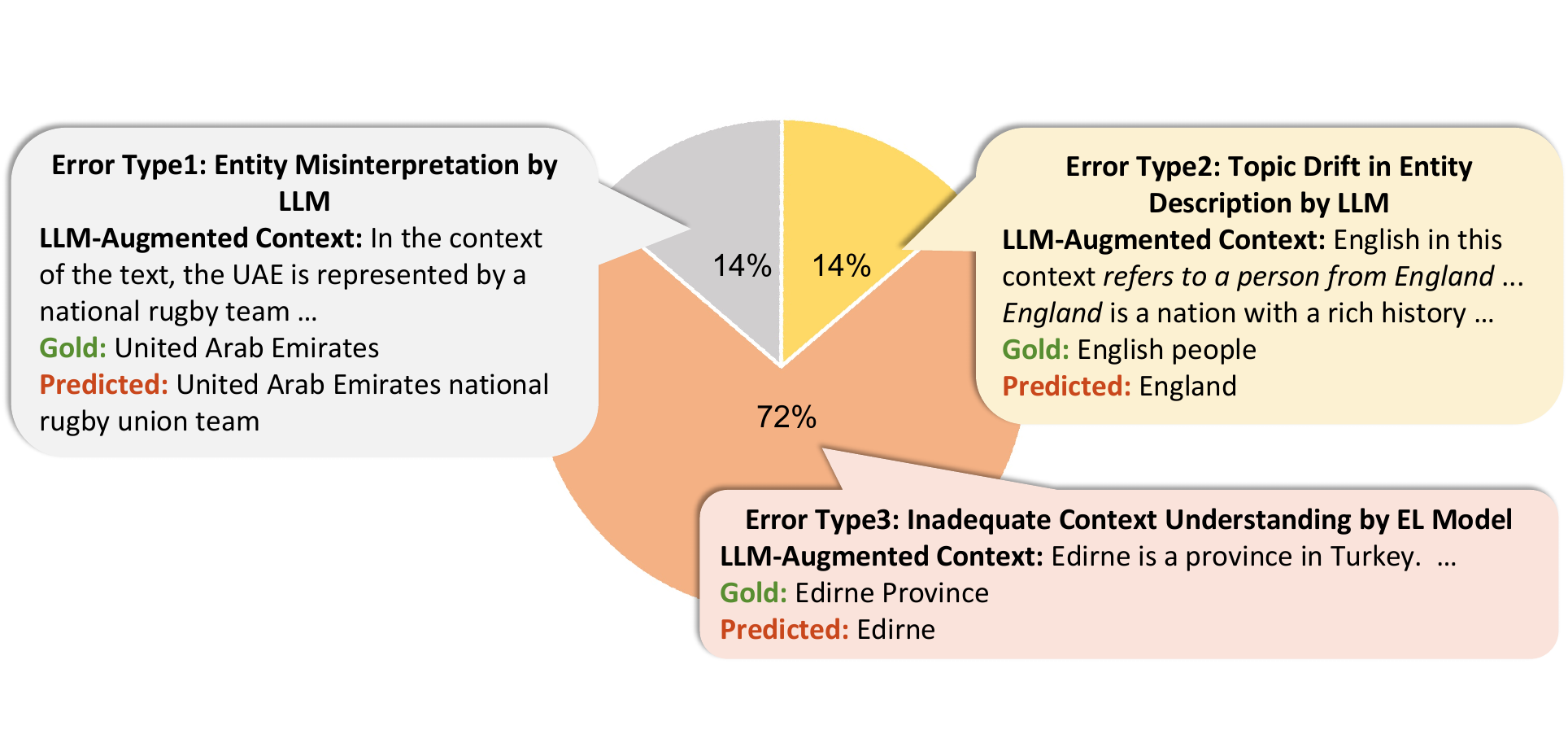}
    \caption{Error analysis of \model.}
    \label{fig:error_analysis}
\end{figure*}

We delve deeper into \model by examining the following two discussion questions.

\subsubsection{Does LLM-Augmented Data Improve EL Performance Over Long-Tail Entities?}
\label{entity_frequencies}

In this section, we examine \model's effectiveness across entities of varying frequencies. EL models typically perform well on high-frequency entities but struggle with low-frequency ones due to limited training data, whereas LLMs encode broader entity knowledge. We hypothesize that \model’s gains stem from improving long-tail entity linking.

To test this, we evaluate \model on MSNBC, AQUAINT, ACE04, and WNED-WIKI. Each entity is assigned a Wikidata5M~\citep{wang2021kepler} PageRank score as its frequency, normalized on a base-10 logarithmic scale and grouped into seven buckets.

Figure~\ref{fig:frequency_bin} shows bucket accuracies: the green line is the baseline ReFinED, while purple lines show $\text{ReFinED}_{\text{FT}}$ trained with LLM-augmented data. Results confirm that \model substantially improves mid- to low-frequency entities ($10^{-6}$–$10^{-2}$) and refines performance on extremely rare ones ($10^{-7}$–$10^{-6}$). This supports our hypothesis that LLM augmentation benefits long-tail entities, while also boosting mid-tail performance by providing cleaner, condensed mention contexts.

\subsubsection{Is There a Better Way to Leverage LLMs for EL?} 

\begin{table}
\caption{\label{rerank}Results of various LLM-for-EL approaches.}
    \scalebox{1.1}{
    \setlength{\tabcolsep}{4pt}
    \begin{tabular}{lcccc}
       \toprule
       \textbf{Method} & \textbf{AIDA} & \textbf{MSNBC} & \textbf{AQUA} & \textbf{ACE04} \\
       \midrule
       BLINK only & $\textbf{82.01}$ & $\textbf{86.23}$ & $\textbf{85.16}$ & $\textbf{86.01}$ \\
       LLM only & $78.37$ & $80.49$ & $73.18$ & $83.27$ \\
       Re-rank-100 & $70.95$ & $82.01$ & $73.18$ & $74.32$ \\
       Re-rank-10 	& $73.24$ & $80.18$ & $73.45$ & $82.88$ \\
       \bottomrule
    \end{tabular}
    }
\end{table}

\model investigates the effective collaboration between LLMs and specialized EL models by LLM context augmentation.
In this section, we explore another possible way to combine LLMs with EL models: using specialized EL models for candidate entity retrieval, and then leveraging LLMs to conduct entity re-ranking.

Considering that many EL models, such as BLINK, operate by first retrieving candidate entities and then re-ranking them, a practical approach to combine LLMs and specialized EL models is to use EL models for candidate entity retrieval and LLMs for entity re-ranking.
We use BLINK's bi-encoder for candidate retrieval and Llama-3-70b-Instruct for re-ranking.
We evaluate two entity re-ranking settings: 

\begin{itemize}
    \item {\bf Re-rank-100: } Extract the top $100$ candidate entities from BLINK's bi-encoder and let the LLM select the final entity.
    \item {\bf Re-rank-10: } Extract the top $10$ candidate entities of BLINK's bi-encoder and augment each candidate with its Wikipedia abstract. The LLM selects the final entity using the abstracts as supplementary information.
\end{itemize}

As shown in Table~\ref{rerank}, using LLMs for entity re-ranking does not improve EL performance.
We observe two primary reasons. 
First, the presence of similar candidate names confuses the LLM. 
Unlike demanding LLMs to directly generate entity names for mentions, asking LLMs to perform re-ranking requires them to discern the subtle distinctions among candidates. 
As highlighted by \citet{peng2023does}, LLMs struggle to understand and distinguish complex contexts, leading to diminished performance.
Secondly, presenting the LLM with multiple candidates often causes it to spread its focus across the entire context rather than concentrating on the specific mention.
This results in the LLM prioritizing information that is distant and unrelated to the mention. 
The suboptimal performance of LLMs used as either direct EL executors or entity re-rankers underscore that \model's context augmentation is the most effective way to integrate LLMs into EL.

\subsection{Error Analysis}  

To gain deeper insight into how \model's context augmentation affects EL performance, we conducted a manual error analysis of failure cases. 
Specifically, we focused on the WNED-WIKI dataset due to its intermediate size and comprehensive content derived from the Wikipedia encyclopedia.  

Upon examining the test outputs of \model \(\times\) \(\text{ReFinED}_{\text{FT}}\) on WNED-WIKI, we observed that \model context augmentation improved $2.67\%$ of the test cases, turning them from incorrect to correct, demonstrating its usefulness in enhancing EL performance. 
However, $1.19\%$ of the test cases regressed, turning from correct to incorrect, indicating that \model can also introduce errors in certain scenarios. 
To better understand these failure cases, we randomly sampled 50 entries from the $1.19\%$ of regressed cases and performed a detailed manual error analysis.  

Our results are illustrated in Figure~\ref{fig:error_analysis}, along with representative examples. 
The failure cases can be categorized into three types:  

\begin{enumerate}
    \item \textbf{Entity Misinterpretation by LLM} ($14\%$):  
    In this category, the LLM incorrectly interprets the target entity and generates an inaccurate description. For example, as shown in Figure~\ref{fig:error_analysis}, while the original context refers to the UAE as the United Arab Emirates, the LLM misinterprets it as the UAE national rugby team, generating a description of the team. This misleads the EL model to link to the incorrect entity ``United Arab Emirates national rugby union team.'' We attribute this error type to the LLM's misinterpretation during context generation.  

    \item \textbf{Topic Drift in LLM Entity Description} ($14\%$):  
    Here, the LLM correctly identifies the target entity but generates context that diverges to describe another related entity. For example, as illustrated in Figure~\ref{fig:error_analysis}, the LLM correctly interprets the mention ``English'' as referring to English people but produces a lengthy description of the nation of England, which is a separate entity. This misleads the EL model to link to ``England'' instead of ``English people.'' This type of error stems from the LLM's inability to maintain focus on the intended entity.  

    \item \textbf{Inadequate Context Understanding by EL Models} ($72\%$):  
    This is the largest category of errors. In these cases, the LLM-generated context is entirely accurate, but the EL model still produces incorrect results. Such failures may arise from either (a) inherent limitations in the EL model's capabilities, or (b) the EL model's inability to fully adapt to the token distribution of the augmented context, even after fine-tuning.  
    The latter suggests that while fine-tuning improves the compatibility of EL models with augmented contexts, there still remain cases where the models fail to adapt effectively.  
\end{enumerate}  


Overall, the analysis shows that LLMs correctly interpret target entities in most cases, as indicated by the combined $86\%$ share of error types 2 and 3, supporting their strength in entity interpretation and potential for EL tasks. Still, error type 1 reveals occasional misinterpretations, while type 2 shows that even accurate interpretations can be undermined by off-topic context. The high proportion of type 3 further highlights the challenge of fully adapting EL models to LLM-augmented contexts.

\section{Conclusion}
\label{sec:conclusion}


In this paper, we proposed \model, a novel framework that augments specialized EL models with LLM-generated context.
We also explored alternative strategies for integrating LLMs into the EL process, including direct EL execution and entity re-ranking.
Compared to these alternatives, \model’s context augmentation strategy provides the most effective use of LLMs for EL, achieving new state-of-the-art results on 5 of 6 benchmarks in the tuning-free setting and all 6 with lightweight EL model fine-tuning. 
We further validated \model's robustness across three LLM backbones and explored ensemble techniques to further boost performance. 
Our analysis shows that LLM context augmentation consistently benefits mid- to long-tail entities, underscoring its strong potential for advancing entity linking.

\begin{acks}
This work is supported by National Natural Science Foundation of China (62476150), Beijing Natural Science Foundation (L243006), and Tsinghua University (Department of Computer Science)-Siemens Ltd., China Joint
Research Center for Industrial Intelligence and Internet of Things (JCIIOT).
The authors also thank the Deng
Feng Fund for supporting this research, and all anonymous reviewers and
meta-reviewers for their valuable feedback.
\end{acks}



\section*{GenAI Usage Disclosure}

In this work, generative AI tools were utilized during the data construction phase.  
Specifically, given a target mention and its original context, we employed off-the-shelf large language models to generate additional contextual information to support entity linking.  
No generative AI tools were used in the writing of this paper, apart from basic spelling and grammar checks.

\bibliographystyle{ACM-Reference-Format}
\bibliography{sample-base}

\end{document}